\documentclass[11pt, a4paper, logo, singlecolumn, copyright]{gensyn}

\usepackage{booktabs}
\usepackage{hyperref}
\usepackage{url}
\usepackage{alertmessage}
\usepackage{xspace}
\usepackage{graphicx}
\usepackage{multirow}
\usepackage{rotating}
\usepackage{tikz,lipsum}
\usepackage[ruled,vlined]{algorithm2e}

\usepackage[colorinlistoftodos,textsize=tiny,textwidth=35pt]{todonotes}
\newcommand{\Comments}{1}
\newcommand{\mynote}[2]{\ifnum\Comments=1\textcolor{#1}{#2}\fi}
\newcommand{\mytodo}[2]{\ifnum\Comments=1\todo[linecolor=#1!80!black,backgroundcolor=#1,bordercolor=#1!80!black]{#2}\fi}
\ifnum\Comments=1
\paperwidth=\dimexpr \paperwidth + 50pt\relax
\oddsidemargin=\dimexpr\oddsidemargin + 25pt\relax
\evensidemargin=\dimexpr\evensidemargin + 25pt\relax
\marginparwidth=\dimexpr \marginparwidth + 25pt\relax
\fi

\usepackage{array}
\newcolumntype{P}{>{\centering\arraybackslash}p{2.5cm}}
\newcolumntype{M}{>{\centering\arraybackslash\footnotesize}m{.78cm}}
\newcolumntype{S}{>{\centering\arraybackslash\tiny}m{2cm}}

\newtcbtheorem[auto counter,number within=section]{obs}%
{Observation}{fonttitle=\bfseries\upshape, fontupper=\slshape,
	arc=0mm, colback=cyan!5!white,colframe=cyan!75!white}{theorem}

\newtcbtheorem[auto counter,number within=section]{ins}%
{Insight}{fonttitle=\bfseries\upshape, fontupper=\slshape,
	arc=0mm, colback=green!5!white,colframe=green!75!white}{theorem}

\usepackage{amsmath,amsfonts,bm}

\def\eqref#1{equation~\ref{#1}}
\def\1{\bm{1}}

\DeclareMathAlphabet{\mathsfit}{\encodingdefault}{\sfdefault}{m}{sl}
\SetMathAlphabet{\mathsfit}{bold}{\encodingdefault}{\sfdefault}{bx}{n}

\usepackage[utf8]{inputenc} % allow utf-8 input
\usepackage[T1]{fontenc}    % use 8-bit T1 fonts
\usepackage{hyperref}       % hyperlinks
\usepackage{url}            % simple URL typesetting
\usepackage{booktabs}       % professional-quality tables
\usepackage{amsfonts}       % blackboard math symbols
\usepackage{nicefrac}       % compact symbols for 1/2, etc.
\usepackage{microtype}      % microtypography
\usepackage{xcolor}         % colors

\usepackage{amsmath}
\usepackage[ruled,vlined]{algorithm2e}
\usepackage{graphicx}
\usepackage{subcaption}

\newcommand{\SAPO}{\textbf{SAPO}}

\begin{document}

\title{Sharing is Caring: Efficient LM Post-Training with Collective RL Experience Sharing}

% \author[1]{Jeffrey Amico}
% \author[1]{Gabriel P.~Andrade}
% \author[1]{John Donaghy}
% \author[1]{Ben Fielding}
% \author[1]{Tristin Forbus}
% \author[1]{Harry Grieve}
% \author[1]{Semih Kara}
% \author[1]{Jari Kolehmainen}
% \author[1]{Yihua Lou}
% \author[1]{Christopher Nies}
% \author[1]{Edward Phillip Flores Nu\~{n}o}
% \author[1]{Diogo Ortega}
% \author[1]{Shikhar Rastogi}
% \author[1]{Jari Kolehmainen}
% \author[1]{Austin Virts}
% \author[1]{Matthew J Wright}
% %\author[1,2]{}
% %\author[2]{}

% \affil[1]{Gensyn}
\author{Gensyn AI Team \\ 
\textbf{Jeffrey Amico, Gabriel Passamani Andrade$^\dagger$, John Donaghy$^\dagger$, Ben Fielding, Tristin Forbus, Harry Grieve, Semih Kara$^\dagger$, Jari Kolehmainen, Yihua Lou$^\dagger$, Christopher Nies, Edward Phillip Flores Nuño, Diogo Ortega, Shikhar Rastogi$^\dagger$, Austin Virts, Matthew J.~Wright} \\
$^\dagger$\small{Primary contributors.}\\
\small{Authors are listed in alphabetical order.}}

% Can leave this option out if you do not wish to add a corresponding author.
\correspondingauthor{gabriel@gensyn.ai~\&~semih@gensyn.ai}

% Remove these if they are not needed 
% \keywords{} 
% \paperurl{}

% Use the internally issued paper ID, if there is one
% \reportnumber{} % Leave blank if n/a 

% Assign your own date to the report. 
% Can comment out if not needed or leave blank if n/a.
% \renewcommand{\today}{} 

\begin{abstract}
	Post-training language models~(LMs) with reinforcement learning~(RL) can enhance their complex reasoning capabilities without supervised fine-tuning, as demonstrated by DeepSeek-R1-Zero~\citep{deepseekai2025deepseekr1incentivizingreasoningcapability}.
    However, effectively utilizing RL for LMs requires significant parallelization to scale-up inference, which introduces non-trivial technical challenges~(e.g.~latency, memory, and reliability) alongside ever-growing financial costs.
    We present \textbf{S}warm s\textbf{A}mpling \textbf{P}olicy \textbf{O}ptimization~(\SAPO), a fully decentralized and asynchronous RL post-training algorithm. 
    \SAPO~is designed for decentralized networks of heterogenous compute nodes, where each node manages its own policy model(s) while ``sharing'' rollouts with others in the network; no explicit assumptions about latency, model homogeneity, or hardware are required and nodes can operate in silo if desired.
    As a result, the algorithm avoids common bottlenecks in scaling RL post-training while also allowing (and even encouraging) new possibilities.
    By sampling rollouts ``shared'' across the network, it~enables ``Aha moments'' to propagate, thereby bootstrapping the learning process. In this paper we show \SAPO~achieved cumulative reward gains of up to $94\%$ in controlled experiments. We also share insights from tests on a network with thousands of nodes contributed by Gensyn community members running the algorithm on diverse hardware and models during an open-source demo.
    % We also share insights gathered from thousands of Gensyn community members who ran the algorithm across a wide range of hardware and models in an open-source demo.
\end{abstract}

\maketitle

\renewcommand\thefootnote{}\footnotetext{%
Acknowledgments: We thank all Gensyn community members who have contributed to our testnet. Your support makes it possible for us to iterate and experiment at unprecedented scales -- we hope you will continue to support us as we work together to democratize AI, do science in the open, and strive to build a future we all deserve.
}
\renewcommand\thefootnote{\arabic{footnote}}

\section{Introduction}\label{sec:intro}

Improving the capabilities of language models~(LMs) after pre-training has become a central goal in AI research. Reinforcement learning~(RL) has emerged as a powerful tool for this purpose, allowing models to improve through trial and error, rather than relying solely on supervised data~\citep{ziegler2020finetuninglanguagemodelshuman,openai2022instructionfollowing,openai2024learningtoreason,deepseekai2025deepseekr1incentivizingreasoningcapability}. Recent efforts to scale RL for LMs have largely focused on distributed systems that orchestrate large GPU clusters that need to keep policy weights synchronized during training~\citep{mistralai2025magistral,wu2025llamarldistributedasynchronousreinforcement,fu2025areallargescaleasynchronousreinforcement}. Although effective, these approaches are expensive, introduce communication bottlenecks, and often require carefully engineered infrastructure to remain stable and efficient.

To address these challenges, we introduce \textbf{Swarm sAmpling Policy Optimization~(SAPO)}.
\SAPO~is a distributed RL algorithm built for decentralized networks of heterogeneous compute nodes. In this setup, which we call \textbf{swarm}, \textit{each node trains its own policy~(or model) while sharing decoded rollouts}~(e.g., in plain text), enabling lightweight exchange of experience. 
This simple mechanism makes the framework independent of model architecture, learning algorithm, and hardware, allowing heterogeneous contributors to participate without synchronization overhead. As a natural consequence, the system behaves like a multi-agent setup, where diverse models and abundant data enhance exploration and improve generalization. In controlled experiments, we observed that \SAPO~delivers higher sample efficiency and stronger task performance---improving cumulative rewards by up to $94\%$ while sidestepping the costs, bottlenecks, and fragility of conventional distributed RL methods.

While \SAPO~can be applied in any RL setting, including the fine-tuning of large LMs (LLMs), in this work we focus on small LMs (SLMs), i.e., LMs with fewer than 10B parameters \citep{vannguyen2024surveysmalllanguagemodels}. We chose SLMs because swarms are most often implemented on local or edge devices, which typically run smaller models rather than large ones. A concrete example is Gensyn’s RLSwarm~\citep{gensyn2025rlswarm}, which allows thousands of heterogeneous SLMs running locally on consumer grade hardware~(e.g., MacBooks) to interact and train collectively. Thanks to contributions from \textbf{\textit{thousands of Gensyn community members}}, we conducted an open-source demo of \SAPO~that produced the empirical insights in~\S\ref{par:training_in_large_swarm}. Overall, the demo showed that collective training with shared experience makes SLMs learn much faster.

\section{Related Work}

\paragraph{Reinforcement Learning for LM Fine-Tuning}
RL has become a central technique for fine-tuning LMs, aligning their behavior with human preferences~\citep{openai2022instructionfollowing,openai2024learningtoreason}, and enabling improvements in factual accuracy~\citep{tian2023finetuning}, code generation~\citep{le2022coderlmasteringcodegeneration} and reasoning beyond what supervised learning alone can achieve~\citep{openai2022instructionfollowing,openai2024learningtoreason,deepseekai2025deepseekr1incentivizingreasoningcapability}. Unlike supervised methods, RL optimizes models through trial-and-error, with RL from human feedback~(RLHF)~\citep{ziegler2020finetuninglanguagemodelshuman,ouyang2022traininglanguagemodelsfollow} and RL with verifiable rewards~(RLVR)~\citep{gao2024designingeffectiverlreward,lambert2025tulu3pushingfrontiers,deepseekai2025deepseekr1incentivizingreasoningcapability} emerging as the dominant paradigms. RLHF trains a reward model on human preference data, while RLVR leverages rule-based, programmatically verifiable reward functions. In both cases, the resulting rewards are used for fine-tuning LMs via policy-gradient algorithms such as Proximal Policy Optimization~(PPO)~\citep{schulman2017proximalpolicyoptimizationalgorithms}. More recently, extensions such as Group Relative Policy Optimization~(GRPO)~\citep{shao2024deepseekmathpushinglimitsmathematical} and its variants~(e.g., DAPO~\citep{yu2025dapoopensourcellmreinforcement}, VAPO~\citep{yue2025vapoefficientreliablereinforcement}) have refined RL objectives to better capture complex reasoning and reduce memory and computation requirements.

\paragraph{Multi-Agent Methods}
Multi-agent methods have become a central focus in LM research, influencing both model architecture and fine-tuning strategies. These approaches are typically organized around three ideas: debate, specialization~(role-playing), and self-improvement~(bootstrapped reasoning). A major line of work is multi-agent debate~\citep{wu2023autogenenablingnextgenllm,du2023improvingfactualityreasoninglanguage,li2024improvingmultiagentdebatesparse, khan2024debatingwithmorepersuasivellms,liang2024encouragingdivergentthinkinglarge}, where multiple LMs independently answer a query, then refine their responses through iterative dialogue. The final output is selected either by voting or by a dedicated verifier, yielding higher quality answers to use for inference or fine-tuning. When a verifier is introduced, debate naturally overlaps with specialization~\citep{subramaniam2025multiagentfinetuningselfimprovement,park2025maporlmultiagentpostcotrainingcollaborative,li2023camelcommunicativeagents,ma2024coevolvingwithotheryou}, in which agents are given defined roles. MALT~\citep{motwani2025maltimprovingreasoningmultiagent} is a prototypical example of specialization, which uses separate agents for generation, verification, and refinement. Alternatively, self-improvement methods emphasize bootstrapping and iterative self-play~\citep{chen2024selfplayfinetuningconvertsweak,zhao2025siriusselfimprovingmultiagentsystems}. For example, SPIN~\citep{chen2024selfplayfinetuningconvertsweak} trains models to iteratively generate responses that approximate human annotations, while simultaneously learning to distinguish between self-generated and human-provided outputs. Adversarial techniques can be integrated into any of these multi-agent strategies (i.e.~debate, specialization, or self-improvement) to deliberately stress models and improve their robustness and safety~\citep{perez2022redteaminglanguagemodels}. Moreover, many of the techniques referenced above use RL to update model parameters~\citep{wu2025llamarldistributedasynchronousreinforcement,ma2024coevolvingwithotheryou,park2025maporlmultiagentpostcotrainingcollaborative,liao2025marftmultiagentreinforcementfinetuning,liu2025llmcollaborationmultiagentreinforcement}.

\paragraph{Comparing SAPO}
Building on RL-based fine-tuning methods like RLHF and RLVR, \SAPO~uses reward-driven trial-and-error to improve LMs. However, unlike traditional approaches, it does not require a single policy to generate all rollouts nor synchronization among multiple policies. 

From a multi-agent perspective, \SAPO~naturally exhibits collaborative behavior with minimal additional computation. Unlike structured multi-agent frameworks, it does not aim to produce specialized nodes or orchestrated collaboration. Nonetheless, by sharing experiences, nodes indirectly benefit from each other’s exploration and reasoning, yielding richer training signals. This positions \SAPO~as a bridge for interpolating between single-agent RL fine-tuning and structured multi-agent frameworks. Through experience sharing \SAPO~accelerates training by capturing many benefits of multi-agent methods, and through RL fine-tuning it encourages individuals to reap those benefits before ultimately passing them on to others in the swarm. Although communication with the swarm and re-encoding sampled rollouts introduces communication and computational overhead, as will be shown in~\S\ref{sec:results}, models trained with the swarm perform better with fewer rounds of training.
Stated another way, \textit{each individuals' additional costs are outweighed by their collective gains.}

\section{Methodology}\label{sec:prelims}

\subsection{The Swarm}\label{subsec:swarm}
Suppose that we have a decentralized network of $N$ nodes~(i.e.~a swarm) that generate and communicate rollouts with one another in discrete time steps $t \in [T]$, where $T > 0$. Each node $n$ has a set of questions~(or tasks) $\mathcal{Q}^n$, and each question $q \in \mathcal{Q}^n$ has a ground-truth solution $y_q \in \mathcal{Y}^n$. The dataset of node $n$ is
\[
\mathcal{D}^n := \{(q, y_q) \mid q \in \mathcal{Q}^n\}.
\]
Importantly, we require that tasks in $\mathcal{D}^n$ are verifiable~(i.e.~their answers can be efficiently and algorithmically checked for correctness), and that rollouts generated by $n$ have the same~(or compatible) modalities as other nodes in the swarm\footnote{An example of a multimodal swarm is one in which some nodes only generate images while others only generate language. In practice, since nodes filter samples from the swarm locally, the assumption about modalities can be omitted and these rollouts in different modalities would simply be ignored when incompatible.}. We denote the metadata\footnote{Made explicit for clarity, but in many swarm settings this metadata is implicit to the swarm being ``joined''.} of $\mathcal{D}^n$ by $\mathcal{M}^n$, which specifies how each task in the dataset can be verified.

Node $n$ also holds a model~(e.g.~an LM) which, in RL terminology, is represented by a \textbf{policy} $\pi^n$ that maps an appropriate input format to answers. Given a question $q\in\mathcal{Q}^n$~(in the appropriate form), we ask node $n$ to generate $L^n$-many answers $$\mathcal{R}^n(q) := \{a^n_1(q),\dots,a^n_{L^n}(q)\},$$
which forms its \textbf{rollout} in response to $q$.

Throughout this paper we assume the model is an SLM and, for simplicity, that questions are presented directly as prompts. Nonetheless, the framework also supports arbitrary prompt generators, allowing for learned control over question/task difficulty and formatting. Furthermore, the dataset, number of generated answers, and sampled rollouts can all vary with time, however we omit the time subscript for notational convenience.

Although not explored in this paper, it is interesting to note that nodes in the swarm do not necessarily need to partake in training and can use any compatible policy; hence, in principle, humans and other non-traditional policies can serve as generators in the swarm.

\subsection{Swarm Sampling Policy Optimization~(SAPO)}\label{subsec:sapo}

Consistent with standard RL post-training, during each round of training, each node subsamples a batch of questions $\mathcal{B}^n\subseteq \mathcal{Q}^n$ and answers them to generate rollouts. In \SAPO, after being generated and before proceeding with reward calculations, each task-rollout data point can be ``shared'' with or be ``sampled'' by other nodes. Stated formally, each node $n$ \textit{broadcasts a subset of the batch of questions along with their metadata, ground-truth answers, and corresponding rollouts}:
$$\mathcal{C}^n(q) := (q,y_q,\mathcal{R}^n(q),\mathcal{M}^n)\text{~for~}q\in\mathcal{S}^n\subseteq \mathcal{B}^n.$$
We emphasize that the rollouts are shared in a decoded format such that individuals in the swarm can emulate these rollouts as if generated by their own policy, e.g.~\textit{individuals can re-encode and compute token-level values as if the rollout was generated by their policy regardless of how unlikely.}

Subsequently, node $n$ constructs a training set $\mathcal{T}^n$ by subsampling $I^n$-many datapoints from its own rollouts and $J^n$-many from those shared in the swarm:
\begin{align*}
\mathcal{T}^n &= \underbrace{\left\{I^n \text{-many samples from } \bigcup_{q\in\mathcal{B}^n} \mathcal{C}^n(q) \right\}}_{\text{self-rollouts}}
\;\cup\;
\underbrace{\left\{J^n \text{-many samples from } \bigcup_{m \neq n,~q\in\mathcal{S}^m} \mathcal{C}^m(q)\right\}}_{\text{external rollouts}}.
\end{align*}
Importantly, individual nodes have full control over the sampling methodology used for choosing between locally-generated or swarm-sampled rollouts---this is an important mechanism for \textit{allowing individuals in the swarm to both tailor and filter through experiences being shared with their policy model}. For example, in the experiments discussed throughout~\S\ref{sec:results}, all nodes first discard rollouts with zero advantage, and then uniformly sample from the remaining swarm rollouts.
% For example, in the experiments discussed throughout~\S\ref{sec:results}, all nodes first filter out any rollouts with zero advantage then do uniform random sampling over the remaining rollouts coming from the swarm. 

After constructing a training set, the node then uses its local reward model $\rho^n$ to compute rewards over $\mathcal{T}^n$, and updates its policy with any policy gradient algorithm, e.g.~PPO or GRPO. This process of constructing training sets from locally-generated and swarm-sampled datapoints repeats for the individual's desired number of rounds. We present a pseudocode summary of \SAPO~in Algorithm~\ref{algo:sapo}. Note that setting $J^n = 0$ reduces node $n$'s training to standard RL fine-tuning.

\setlength{\algotitleheightrule}{0.8pt}
\setlength{\algotitleheightrule}{0.8pt}
\setlength{\interspacetitleruled}{2ex} % space around title
\setlength{\algotitleheightrule}{0.8pt}
\setlength{\algotitleheightrule}{0.8pt}
\SetAlCapSkip{1.2em} % space after caption
\renewcommand{\baselinestretch}{1.2} % extra line spacing
\begin{algorithm}[htbp!]
\DontPrintSemicolon
\SetAlgoNlRelativeSize{-1}
\SetAlgoNlRelativeSize{-1}
\SetKwInput{KwIn}{Input}
\SetKwInput{KwOut}{Output}

\KwIn{For each $n\in[N]$: dataset~$\mathcal{D}^n$, metadata~$\mathcal{M}^n$, policy~$\pi^n$, reward model~$\rho^n$, policy update algorithm, number of local samples $I^n$, number of external samples $J^n$}
\KwOut{Updated policy parameters}

\For{each round $t$}{
    \For(\tcp*[f]{Can be fully decentralized and run in parallel}){each node $n$}{
        \tcp{Sample questions}
        $\mathcal{B}^n \gets \text{SampleBatch}(\mathcal{Q}^n)$ \;
        \tcp{Generate rollouts}
        \For{each $q \in \mathcal{B}^n$}{
            $\mathcal{R}^n(q) \gets \{ a^n_{1}(q), \dots, a^n_{L^n}(q) \}$ \;
        }
        
        \tcp{Share rollouts and associated data}
        $\mathcal{S}^n \gets \text{SelectSubset}(\mathcal{B}^n)$ \;
        \text{Communicate}($\{\mathcal{C}^n(q)\mid q\in\mathcal{S}^n\}$)
        
        \tcp{Assemble training set}
        $\mathcal{T}^n \gets \text{SampleSelf}(\{\mathcal{C}^n(q) \mid q\in\mathcal{B}^n\}, I^n)$ \\
        $\qquad\quad \hookrightarrow ~\text{SampleExternal}(\cup_{m\neq n}\{\mathcal{C}^m(q) \mid q\in\mathcal{S}^m\}, J^n)$ \;
        
        \tcp{Policy update}
        $\pi^n \gets \text{PolicyGradient}(\pi^n, \rho^n, \mathcal{T}^n)$ \;
    }
}
\caption{\SAPO}
\label{algo:sapo}
\end{algorithm}

\section{Controlled Experiment Setup}\label{sec:experiment_setup}
In this section, we describe the setup of our controlled experiments. We used a swarm of eight Qwen2.5 models~\citep{qwen2.5}, each with 0.5B parameters, implemented the training process using PyTorch and ran the models within Docker containers. Docker Compose scripts orchestrated the containers, enabling scalable deployment. We managed multi-GPU parallelism with PyTorch's \texttt{distributed} package, assigning one GPU per agent and allowing NCCL to facilitate their communication in a swarm.

\subsection{Dataset}\label{subsec:data}
% For our experiments we used the ReasoningGYM dataset~\citep{stojanovski2025reasoninggymreasoningenvironments}, which procedurally generates multi-domain~(e.g., algebra, logic, or graph problems) data at run time via domain-specific generators that produce fresh problem instances each time they are called. 
We conducted our experiments using the ReasoningGYM dataset~\citep{stojanovski2025reasoninggymreasoningenvironments}. This dataset generates problems on demand in domains such as algebra, logic, and graph reasoning. Each time a problem is requested, a domain-specific generator produces a fresh instance. Hence, ReasoningGYM can technically yield an unlimited stream of diverse training and evaluation tasks, with adjustable size, structure, and difficulty. Each domain-specific generator is also paired with a programmatic verifier, enabling reliable ``off-the-shelf'' correctness checks.

From amongst the catalog of tasks available in reasoning gym\footnote{The full gallery of ReasoningGYM datasets can be found at the following link: \href{https://github.com/open-thought/reasoning-gym/blob/main/GALLERY.md}{https://github.com/open-thought/reasoning-gym/blob/main/GALLERY.md}}, we selected the following specialties: 
\begin{itemize}
    \item \texttt{base\_conversion}: converting numbers between different bases;
    \item \texttt{basic\_arithmetic}: performing elementary arithmetic operations;
    \item \texttt{arc\_1d}: abstract reasoning over one-dimensional sequences~(a simplified version of the ARC benchmark);
    \item \texttt{bf}: tasks involving Brainf*ck programs or similar algorithmic reasoning;
    \item \texttt{propositional\_logic}: solving propositional logic questions;
    \item \texttt{fraction\_simplification}: simplifying fractions as much as possible;
    \item \texttt{decimal\_arithmetic}: enforcing proper operator precedence during arithmetic with decimal contraints;
    \item \texttt{calendar\_arithmetic}: puzzle solving on word problems involving calendar dates;
    \item \texttt{binary\_matrix}: abstract reasoning on binary square matrices.
\end{itemize}
This selection ensures evaluation across a diverse set of reasoning tasks spanning symbolic, numeric, and abstract domains. At each training round, every agent randomly samples a set of specialties~(with replacement) from the above list and receives one ReasoningGYM question per specialty. For each of its own questions, each agent generates $8$ completions, resulting in a rollout of 8 entries per question~(i.e.~$L^n=8$ for all $n\in[8]$). We emphasize that, in the experiments, agents are generalists; i.e., they received questions from all specialties with equal probability.
% An interesting next step for evaluating \SAPO~would be to study the effect of specialization~(or greater heterogeneity).

\subsection{Policy Update}\label{subsec:local_alg}
We used GRPO to update each node's policy. In initial experiments, as identified in DAPO~\citep{yu2025dapoopensourcellmreinforcement}, we found training to be more efficient without the KL-divergence penalty, so we set its weight to zero. Similarly, for clipping we used asymmetric thresholds with $\epsilon_{\text{low}} = 0.2$~(lower ratio bound) and $\epsilon_{\text{high}} = 0.28$~(upper ratio bound). Training ran for $2000$ rounds and used the Adam optimizer with the default hyperparameters~(e.g.,~learning rate $0.001$).

\subsection{Reward Model}\label{subsec:rewards}
For each node and task, we used the flexible, rule-based verifiers provided by ReasoningGYM. If the task-specific verifier was able to parse the correct answer from a completion, then it assigned a reward of $1$~\footnote{There are exceptions in specific verifiers where completions can receive partial credit for edge cases.} otherwise $0$. 
% that completion receives a reward of $1$~\footnote{There are occasional exceptions in specific verifiers where completions can receive partial credit for edge cases.} otherwise $0$. 

Interestingly, in our early experiments we added a formatting reward, but soon removed it. Experience sharing in \SAPO~made it unnecessary because knowledge about the correct formatting (expected by ReasoningGYM’s verifiers) spread throughout the swarm almost immediately without needing an explicit formatting reward signal.

% Interestingly, during initial experimentation we also included a formatting reward but soon removed it after finding that the experience sharing amongst nodes in \SAPO~made it obsolete--knowledge about the correct formatting expected by ReasoningGYM's verifiers was being propogated throughout the swarm almost immediately without needing to explicitly include a formatting-related signal.

\subsection{GenRL}
We used GenRL~\citep{gensyn2025genrl}, the backend for Gensyn’s RLSwarm platform, to perform our experiments. GenRL is a decentralized, modular framework designed for scalable, multi-agent, multi-stage reinforcement learning. Unlike centralised systems, it supports peer-to-peer coordination and communication, giving full control over system architecture. Notably, GenRL integrates seamlessly with ReasoningGYM, offering out-of-the-box access to over 100 procedurally generated, verifiable reasoning tasks.

\section{Controlled Experiment Results}\label{sec:results}

To evaluate the efficacy of \SAPO, we trained the swarm described in~\S\ref{sec:experiment_setup} \textit{with varying degrees of experience sharing}. The \textit{baseline is no sharing}, which corresponds to standard RL fine-tuning~(\S\ref{sec:prelims}). To keep conditions comparable, each agent was assigned 8 tasks~(questions) per round, ensuring the total number of training samples was fixed across all setups.

In the baseline, each agent samples specialties uniformly, receives one question per specialty, generates $8$ completions per question, and updates its policy using only~(and all of) its own rollouts. In \SAPO~with $I$ local~/~$J$ external rollouts (where $I,J>0$ and $I+J=8$), each agent samples $I$ specialties, receives one question per specialty, and generates 8 completions per question (number of completions per question is fixed across all experiments). Each agent then shares all of its rollouts with the swarm. From the shared pool, it removes rollouts with zero advantage, samples $J$ rollouts from the remaining ones, and combines them with its own local rollouts to update its policy. Note that \SAPO~gives agents the flexibility to \textit{subsample from a larger pool and remove the uninformative, $0$-advantage samples}, which the baseline cannot do. 

We evaluate four configurations in total: $8$ local~/~$0$ external~(baseline), $6$ local~/~$2$ external, $4$ local~/~$4$ external, and $2$ local~/~$6$ external. For each configuration, we examine the reward trajectories produced during training.

\begin{figure}[ht]
    \centering
    
    % Row 1
    \begin{subfigure}{0.49\textwidth}
        \centering
        \includegraphics[width=\linewidth]{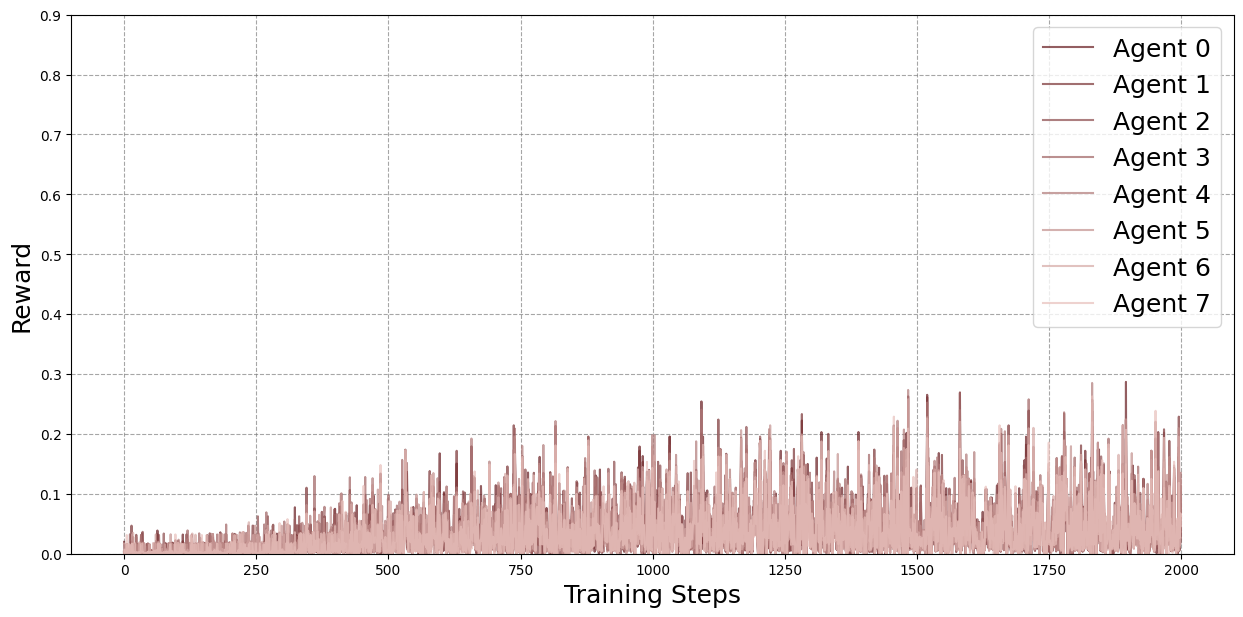}
        \caption{Baseline case, i.e.~8 local / 0 external rollouts.}
        \label{fig:raw_rewards80}
    \end{subfigure}
    \hfill
    \begin{subfigure}{0.49\textwidth}
        \centering
        \includegraphics[width=\linewidth]{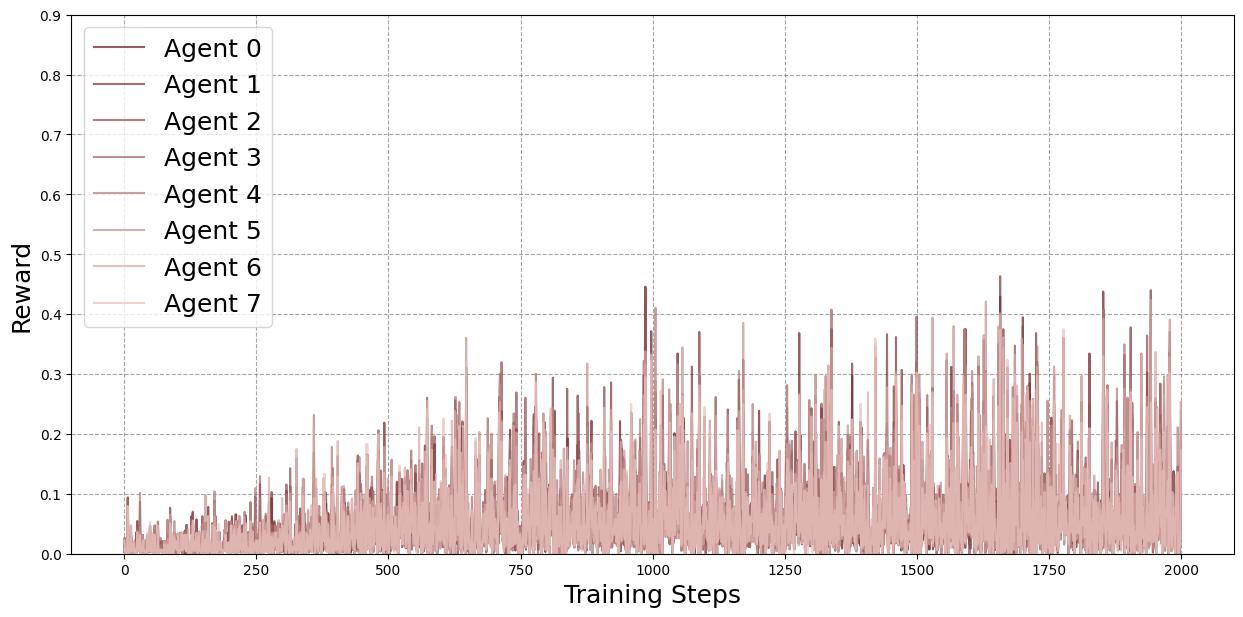}
        \caption{6 local / 2 external rollouts.}
        \label{fig:raw_rewards62}
    \end{subfigure}
    
    % Row 2
    \begin{subfigure}{0.49\textwidth}
        \centering
        \includegraphics[width=\linewidth]{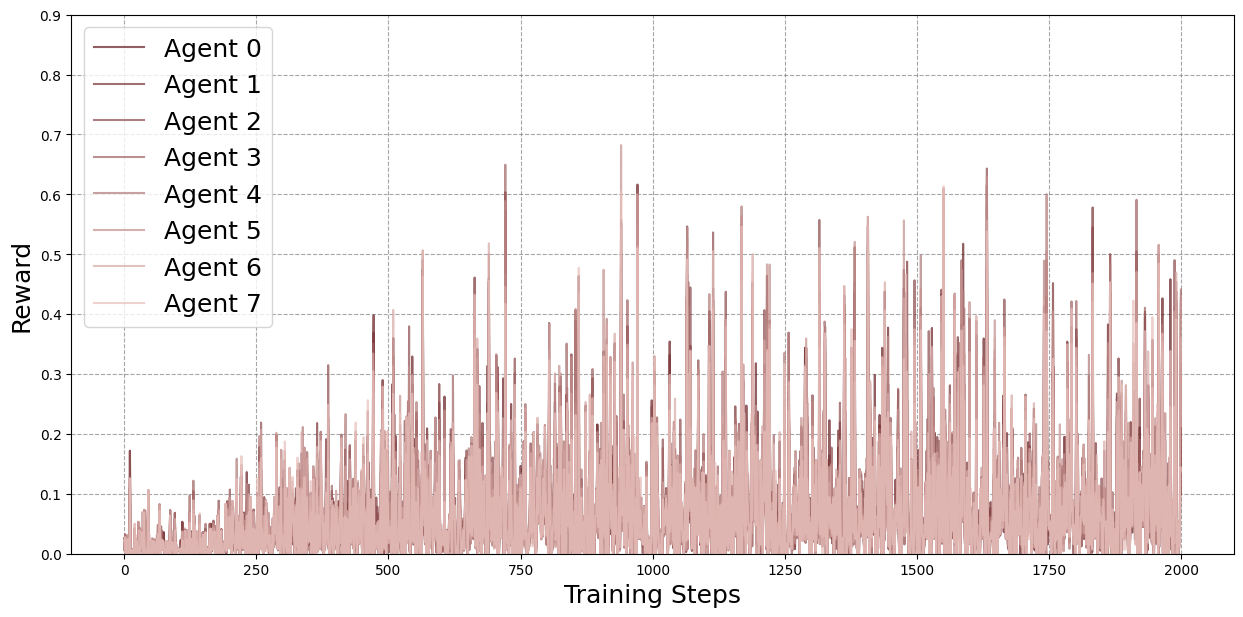}
        \caption{4 local / 4 external rollouts.}
        \label{fig:raw_rewards44}
    \end{subfigure}
    \hfill
    \begin{subfigure}{0.49\textwidth}
        \centering
        \includegraphics[width=\linewidth]{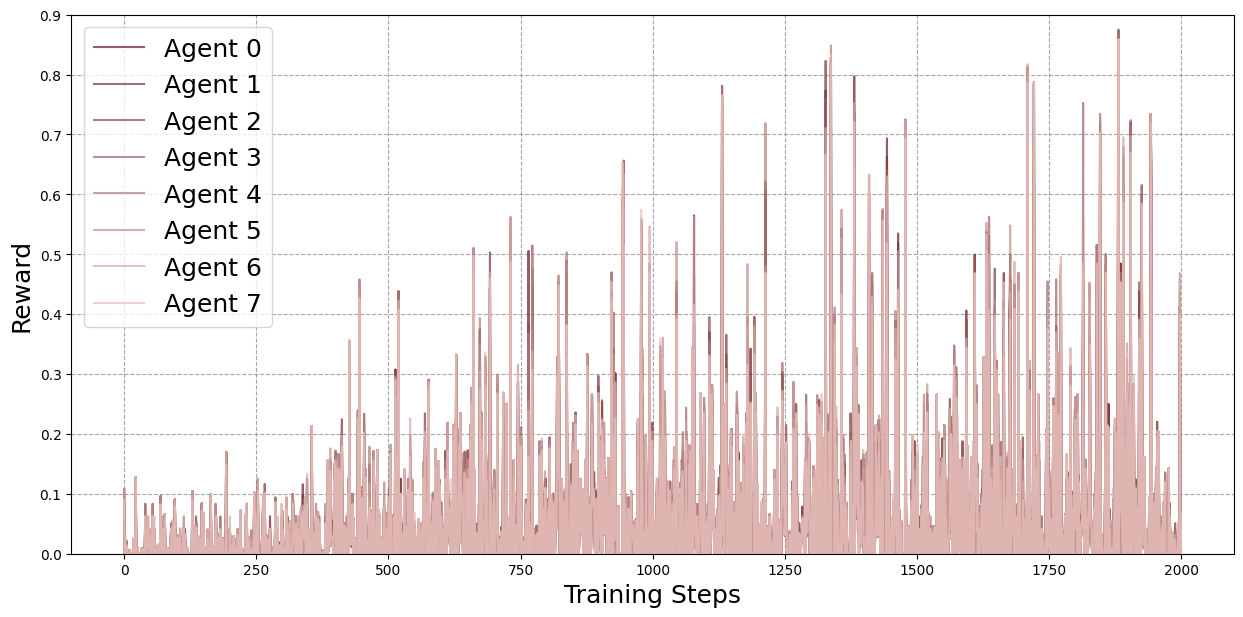}
        \caption{2 local / 6 external rollouts.}
        \label{fig:raw_rewards26}
    \end{subfigure}

    \caption{Rewards obtained by all agents are shown for each configuration. Increasing the number of external rollouts raises peak rewards, but the highest overall reward accumulation occurs in the $4$ local~/~$4$ external setup, yielding a $\%94$ improvement over the baseline.}
    \label{fig:raw_rewards}
\end{figure}

Figure~\ref{fig:raw_rewards} shows the rewards obtained by each agent across training rounds, with each subplot corresponding to a different configuration. Both the $4$ local~/~$4$ external and $2$ local~/~$6$ external schemes achieve the highest peak rewards, clearly outperforming the no-sharing baseline. Among all setups, the $4$ local~/~$4$ external configuration achieves the largest total reward accumulated across agents and rounds~($1093.31$), followed by $2$ local~/~$6$ external~($945.87$) and $6$ local~/~$2$ external~($854.43$). By comparison, the baseline yields only $561.79$. Overall, the $4$ local / $4$ external scheme delivers the strongest performance, with a $94\%$ improvement over the baseline.

\begin{figure}[ht]
    \centering
    
    \includegraphics[width=\linewidth]{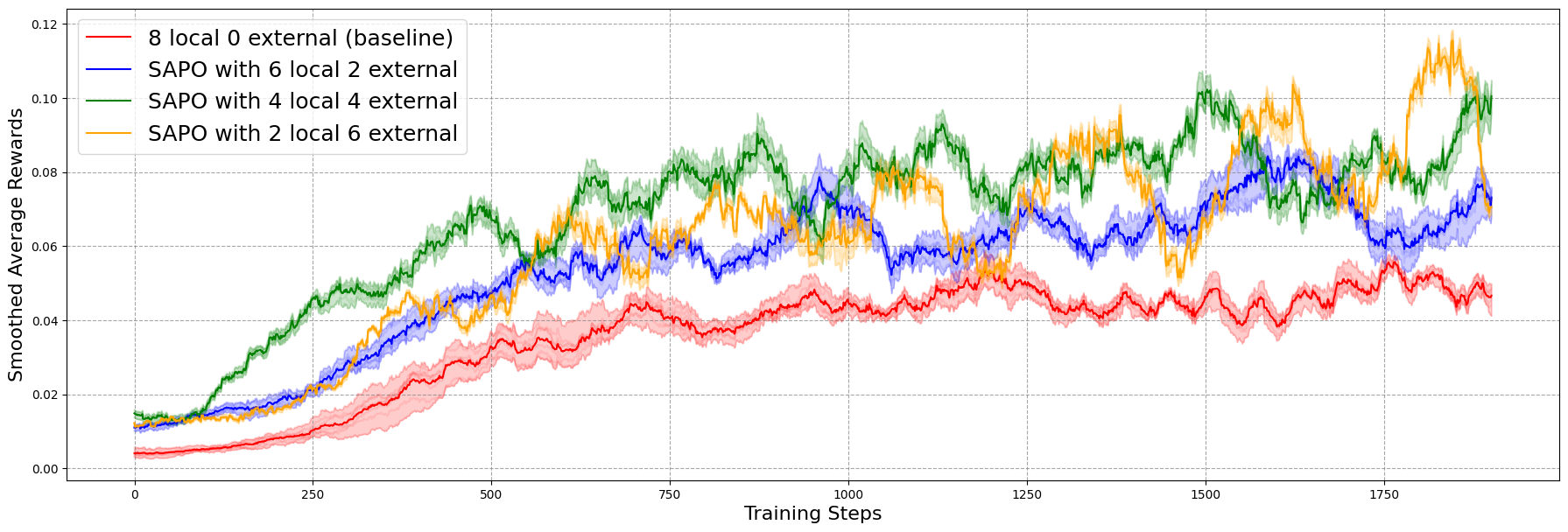}
    
    \caption{Average agent rewards for each configuration across training, smoothed with a moving average~(window size 100). The 4 local / 4 external configuration consistently outperforms the baseline and, in nearly all rounds, also exceeds the 6 local / 2 external configuration in expected average reward. The 4 local / 4 external configuration also surpasses the 2 local / 6 external setup for most rounds, though the difference is smaller compared to the other cases.}
    \label{fig:smooth_rewards}
\end{figure}

To gain further insight, we examine the agent-averaged rewards for each configuration. In addition, we apply a moving average with a window of $100$~samples to smooth the curves. Since the policy parameters change slower than individual training steps, the moving average effectively averages rewards as if the policy were frozen. As a result, the smoothed curve serves as a reasonable estimate of the expected average reward across tasks. The results are shown in Figure~\ref{fig:smooth_rewards}, with confidence intervals given by the minimum and maximum across agents.

Figure~\ref{fig:smooth_rewards} illustrates that the $4$ local~/~$4$ external configuration consistently achieves higher expected average reward than the baseline, and in nearly all training rounds, it also outperforms the $6$ local~/~$2$ external configuration. Compared to the $2$ local~/~$6$ external setup, the $4$ local~/~$4$ external configuration again performs better for most rounds, although the difference is smaller than in the other cases. \textit{These results highlight the benefit of experience sharing: once one agent has an ``Aha moment,'' it can spread through the swarm and lift overall performance.} Notably, this effect appears even without giving agents specific roles or different models; adding more heterogeneity could make the swarm effect even stronger and suggests a promising line of future work for~\SAPO.

On the other hand, since the $4$ local~/~$4$ external configuration outperforms the $2$ local~/~$6$ external case, we find that \textit{relying too heavily on external rollouts can actually hinder performance}. Notice also that the baseline shows much lower variation, and the level of oscillation increases as the proportion of external rollouts grows. The $2$ local~/~$6$ external setup, in particular, shows strong oscillations as training progresses. We interpret this as being due to two interesting network effects:~(i)~When high-performing agents overly rely on external rollouts, their progress can be adversely effected by the answers of worse-performing agents. 
(ii)~When agents draw many rollouts from the swarm but collectively contribute too few, the quality of the shared pool diminishes.
% (ii)~When the number of external rollouts being sampled from the swarm is large and not enough ``contributions'' are being made to the swarm, there is a higher likelihood that swarm-samples are of lower quality.
Taken together, these effects lead to steep learning and forgetting behavior that explains the oscillatory pattern. 
Finally, note that the moving average window smooths out task-level idiosyncrasy~(see Figure~\ref{fig:raw_rewards}), so these oscillations reflect meaningful large-scale training dynamics rather than task related randomness.

% \begin{figure}[ht]
%     \centering
    
%     % Top figure
%     \begin{subfigure}{0.9\textwidth}  % adjust width as needed
%         \centering
%         \includegraphics[width=\linewidth]{figures/SAPO.png}
%         \caption{Smoothed average rewards.}
%         \label{fig:SAPO}
%     \end{subfigure}
    
%     \vspace{0.5cm} % optional vertical spacing
    
%     % Bottom figure
%     \begin{subfigure}{0.9\textwidth}
%         \centering
%         \includegraphics[width=\linewidth]{figures/SAPO_no_2loc6ext.png}
%         \caption{The 2 local / 6 external case is omitted for easier comparison.}
%         \label{fig:SAPO_no_2loc6ext}
%     \end{subfigure}

%     \caption{Average agent rewards for each configuration across training, smoothed with a moving average~(window size 100). Figure~\ref{fig:SAPO} shows all configurations, while Figure~\ref{fig:SAPO_no_2loc6ext} omits the 2 local / 6 external case for easier comparison. The 4 local / 4 external configuration consistently outperforms the baseline and, in nearly all rounds, also exceeds the 6 local / 2 external configuration in expected average reward. The 4 local / 4 external configuration also surpasses the 2 local / 6 external setup for most rounds, though the margin is smaller compared to the other cases.}
%     \label{fig:smooth_rewards}
% \end{figure}

\section{Training in a Large Swarm: Insights from an Open-Source Demo}
\label{par:training_in_large_swarm}
To evaluate \SAPO~in more realistic and heterogeneous conditions, we collected data from a large-scale open source demo in which thousands of Gensyn community members contributed training runs across diverse hardware and model configurations. Each participating node had a unique peer identifier associated with metadata such as the model type being trained and, after each round, nodes participated in the following exchange with a ``judge'' we controlled:~(i)~the node requested an evaluation, (ii)~the judge randomly sampled a question from one of the ReasoningGYM tasks introduced in~\S\ref{subsec:data} and sent it to the node, (iii)~the node generated an answer to the question~(i.e., pass@$1$) and submitted it to the judge, (iv)~the judge scored the answer with the appropriate ReasoningGYM verifier.

By analyzing the result of these judge evaluations associated with unique peer identifiers, we were able to compare models trained collaboratively in the swarm against counterparts trained in isolation. Our findings show that swarm-based training can yield measurable gains, but the effect is model dependent. 

\begin{figure}[htbp!]
	\centering
	\includegraphics[width=\textwidth]{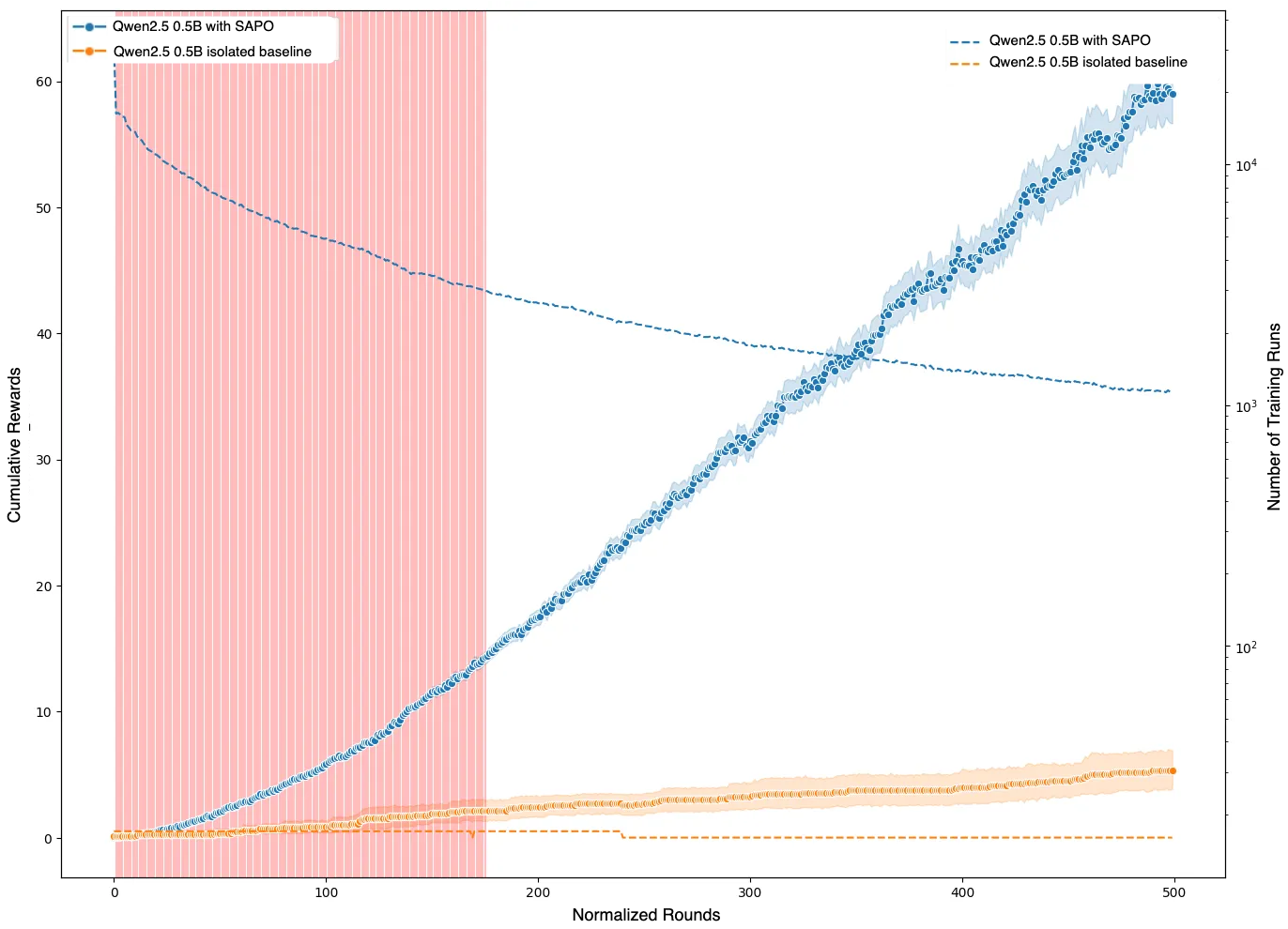}
	\caption{Shown in red are the regions where the adjusted p-value is greater than $0.05$.  After a certain number of rounds, in this case approximately $175$, the performance per round of the models in the swarm significantly exceeds that of the model trained in isolation.}
	\label{fig:demo}
\end{figure}

For the Qwen2.5 models with 0.5B parameters, swarm participation consistently led to improved cumulative performance over time compared to isolated training. As shown in Figure~\ref{fig:demo}, statistical testing confirms that after roughly $175$ normalized rounds\footnote{The large-scale demo's swarm was an ephemeral environment where nodes came and went or occasionally stopped then restarted. Hence we normalize rounds based on how many total rounds individuals participated in.}, the swarm-trained variant outperforms its isolated counterpart. Interestingly, by contrast, stronger models such as Qwen3 with 0.6B parameters achieved similar performance in and out of the swarm, suggesting that \SAPO’s benefits are most pronounced for mid-capacity models that can actively ``absorb'' and propagate diverse rollouts. 

It is important to note that, in this demo, models selected rollouts from the swarm using straightforward uniform random sampling without any filtering. This caused rollouts without useful reward signals to be overrepresented in the swarm. We hypothesize that, with better sampling strategies, more performant models could also benefit from participating in the swarm under \SAPO.

% It is important to note that models in this demo sampled rollouts from the swarm based on a top-k strategy, wherein they sampled rollouts having the highest total rewards across completions without any filtering; as a result, relatively easy questions were disproportionately favored by models in the swarm. We hypothesize that, with better sampling strategies, more performant models would also benefit from participating in the swarm while using \SAPO.

\section{Conclusion}\label{sec:conclusion}
In this paper we introduced \SAPO, a fully decentralized and asynchronous RL post-training algorithm. Unlike many centralized approaches, \SAPO~allows heterogeneous nodes to manage their own models while sharing rollouts, enabling learning to spread across the swarm without assumptions about latency, homogeneity, or hardware. Our experiments, using the ReasoningGYM dataset, show that balanced experience sharing~($4$ local~/~$4$ external) nearly doubles performance over the no-sharing baseline. However, excessive reliance on external rollouts can destabilize learning, causing steep learning and forgetting behavior. Overall, \SAPO~turns experience sharing into a core advantage, offering a scalable and practical path towards augmenting the reasoning capabilities of models through collaborative post-training.

\paragraph{Future Directions}
As discussed in~\S\ref{subsec:data} and~\S\ref{sec:results}, a natural next step is to evaluate \SAPO~under greater heterogeneity---for example, with specialized tasks or different base models. We presented some preliminary results in~\S\ref{par:training_in_large_swarm} where several different base models were allowed to participate in a large swarm, but a more systematic study is needed. Taking these explorations of heterogeneity to their natural extreme, as noted in~\S\ref{sec:prelims}, it would be interesting to explore the effect of unconventional, non-trained policies~(e.g., humans) within the swarm when proper incentive mechanisms are put in place to encourage earnest contributions.

Stability remains an important open question: heavy reliance on external rollouts often causes oscillations and forgetting. Hybrid approaches that integrate \SAPO~with reward-guided sharing, RLHF, or generative verifiers~\citep{zhang2025generativeverifiersrewardmodeling} may help resolve this. Complimentary to these hybrid approaches, especially in large swarm settings where trust cannot be assumed, a promising direction is to develop meta-strategies for adaptively balancing local vs.~shared rollouts, or for strategically filtering swarm samples. 

% Another key direction is addressing the oscillations and forgetting we observed when training relies too heavily on external rollouts, particularly for long-term stability. Hybrid approaches that combine \SAPO~with reward guided experience sharing and RLHF or generative verifiers~\citep{zhang2025generativeverifiersrewardmodeling} may help stabilize training and, in doing so, could unlock the full potential of experience sharing. Complimentary to these hybrid approaches, especially in large swarm settings where trust cannot be assumed, exploring meta-strategies for dynamically choosing the ratio of locally-generated to~swarm-sampled rollouts that nodes should utilize in a round or how to better filter rollouts being sampled from the swarm is a direction that could dramatically improve training outcomes.

Finally, although our focus in this paper was on language models, \SAPO~is agnostic to data modality and can be applied quite generally; multi-modal applications of \SAPO~suggest intriguing directions that are quite unintuitive to imagine in traditional single-agent learning. For example, in image-based~(or any data with a notion of ``aesthetic'') swarms, individual nodes can define local reward mechanisms that codify a personal sense of ``taste'' which can induce a virtuous cycle if deemed ``good'' by other nodes' reward mechanisms since it may indirectly influence the style of images produced by other models in the swarm\footnote{In GenRL~\citep{gensyn2025genrl} we provide an example of a text-to-image swarm where some nodes assign rewards based only on aesthetics whereas others assign rewards based only on CLIPScore~\citep{hessel2022clipscorereferencefreeevaluationmetric}. The resulting policies generate images that satisfy both types of rewards.}. More broadly, multi-agent experience sharing algorithms such as \SAPO~allow us to explore novel learning paradigms designed to leverage unique self-organizing feedback loops between heterogenous models communicating with one another.

\bibliography{references}

\begin{thebibliography}{38}
\providecommand{\natexlab}[1]{#1}
\providecommand{\url}[1]{\texttt{#1}}
\expandafter\ifx\csname urlstyle\endcsname\relax
  \providecommand{\doi}[1]{doi: #1}\else
  \providecommand{\doi}{doi: \begingroup \urlstyle{rm}\Url}\fi

\bibitem[Chen et~al.(2024)Chen, Deng, Yuan, Ji, and Gu]{chen2024selfplayfinetuningconvertsweak}
Zixiang Chen, Yihe Deng, Huizhuo Yuan, Kaixuan Ji, and Quanquan Gu.
\newblock Self-play fine-tuning converts weak language models to strong language models, 2024.
\newblock URL \url{https://arxiv.org/abs/2401.01335}.

\bibitem[DeepSeek-AI et~al.(2025)DeepSeek-AI, Guo, Yang, Zhang, Song, Zhang, Xu, Zhu, Ma, Wang, Bi, Zhang, Yu, Wu, Wu, Gou, Shao, Li, Gao, Liu, Xue, Wang, Wu, Feng, Lu, Zhao, Deng, Zhang, Ruan, Dai, Chen, Ji, Li, Lin, Dai, Luo, Hao, Chen, Li, Zhang, Bao, Xu, Wang, Ding, Xin, Gao, Qu, Li, Guo, Li, Wang, Chen, Yuan, Qiu, Li, Cai, Ni, Liang, Chen, Dong, Hu, Gao, Guan, Huang, Yu, Wang, Zhang, Zhao, Wang, Zhang, Xu, Xia, Zhang, Zhang, Tang, Li, Wang, Li, Tian, Huang, Zhang, Wang, Chen, Du, Ge, Zhang, Pan, Wang, Chen, Jin, Chen, Lu, Zhou, Chen, Ye, Wang, Yu, Zhou, Pan, Li, Zhou, Wu, Ye, Yun, Pei, Sun, Wang, Zeng, Zhao, Liu, Liang, Gao, Yu, Zhang, Xiao, An, Liu, Wang, Chen, Nie, Cheng, Liu, Xie, Liu, Yang, Li, Su, Lin, Li, Jin, Shen, Chen, Sun, Wang, Song, Zhou, Wang, Shan, Li, Wang, Wei, Zhang, Xu, Li, Zhao, Sun, Wang, Yu, Zhang, Shi, Xiong, He, Piao, Wang, Tan, Ma, Liu, Guo, Ou, Wang, Gong, Zou, He, Xiong, Luo, You, Liu, Zhou, Zhu, Xu, Huang, Li, Zheng, Zhu, Ma, Tang, Zha, Yan, Ren, Ren, Sha, Fu, Xu, Xie, Zhang,
  Hao, Ma, Yan, Wu, Gu, Zhu, Liu, Li, Xie, Song, Pan, Huang, Xu, Zhang, and Zhang]{deepseekai2025deepseekr1incentivizingreasoningcapability}
DeepSeek-AI, Daya Guo, Dejian Yang, Haowei Zhang, Junxiao Song, Ruoyu Zhang, Runxin Xu, Qihao Zhu, Shirong Ma, Peiyi Wang, Xiao Bi, Xiaokang Zhang, Xingkai Yu, Yu~Wu, Z.~F. Wu, Zhibin Gou, Zhihong Shao, Zhuoshu Li, Ziyi Gao, Aixin Liu, Bing Xue, Bingxuan Wang, Bochao Wu, Bei Feng, Chengda Lu, Chenggang Zhao, Chengqi Deng, Chenyu Zhang, Chong Ruan, Damai Dai, Deli Chen, Dongjie Ji, Erhang Li, Fangyun Lin, Fucong Dai, Fuli Luo, Guangbo Hao, Guanting Chen, Guowei Li, H.~Zhang, Han Bao, Hanwei Xu, Haocheng Wang, Honghui Ding, Huajian Xin, Huazuo Gao, Hui Qu, Hui Li, Jianzhong Guo, Jiashi Li, Jiawei Wang, Jingchang Chen, Jingyang Yuan, Junjie Qiu, Junlong Li, J.~L. Cai, Jiaqi Ni, Jian Liang, Jin Chen, Kai Dong, Kai Hu, Kaige Gao, Kang Guan, Kexin Huang, Kuai Yu, Lean Wang, Lecong Zhang, Liang Zhao, Litong Wang, Liyue Zhang, Lei Xu, Leyi Xia, Mingchuan Zhang, Minghua Zhang, Minghui Tang, Meng Li, Miaojun Wang, Mingming Li, Ning Tian, Panpan Huang, Peng Zhang, Qiancheng Wang, Qinyu Chen, Qiushi Du, Ruiqi Ge, Ruisong
  Zhang, Ruizhe Pan, Runji Wang, R.~J. Chen, R.~L. Jin, Ruyi Chen, Shanghao Lu, Shangyan Zhou, Shanhuang Chen, Shengfeng Ye, Shiyu Wang, Shuiping Yu, Shunfeng Zhou, Shuting Pan, S.~S. Li, Shuang Zhou, Shaoqing Wu, Shengfeng Ye, Tao Yun, Tian Pei, Tianyu Sun, T.~Wang, Wangding Zeng, Wanjia Zhao, Wen Liu, Wenfeng Liang, Wenjun Gao, Wenqin Yu, Wentao Zhang, W.~L. Xiao, Wei An, Xiaodong Liu, Xiaohan Wang, Xiaokang Chen, Xiaotao Nie, Xin Cheng, Xin Liu, Xin Xie, Xingchao Liu, Xinyu Yang, Xinyuan Li, Xuecheng Su, Xuheng Lin, X.~Q. Li, Xiangyue Jin, Xiaojin Shen, Xiaosha Chen, Xiaowen Sun, Xiaoxiang Wang, Xinnan Song, Xinyi Zhou, Xianzu Wang, Xinxia Shan, Y.~K. Li, Y.~Q. Wang, Y.~X. Wei, Yang Zhang, Yanhong Xu, Yao Li, Yao Zhao, Yaofeng Sun, Yaohui Wang, Yi~Yu, Yichao Zhang, Yifan Shi, Yiliang Xiong, Ying He, Yishi Piao, Yisong Wang, Yixuan Tan, Yiyang Ma, Yiyuan Liu, Yongqiang Guo, Yuan Ou, Yuduan Wang, Yue Gong, Yuheng Zou, Yujia He, Yunfan Xiong, Yuxiang Luo, Yuxiang You, Yuxuan Liu, Yuyang Zhou, Y.~X. Zhu,
  Yanhong Xu, Yanping Huang, Yaohui Li, Yi~Zheng, Yuchen Zhu, Yunxian Ma, Ying Tang, Yukun Zha, Yuting Yan, Z.~Z. Ren, Zehui Ren, Zhangli Sha, Zhe Fu, Zhean Xu, Zhenda Xie, Zhengyan Zhang, Zhewen Hao, Zhicheng Ma, Zhigang Yan, Zhiyu Wu, Zihui Gu, Zijia Zhu, Zijun Liu, Zilin Li, Ziwei Xie, Ziyang Song, Zizheng Pan, Zhen Huang, Zhipeng Xu, Zhongyu Zhang, and Zhen Zhang.
\newblock Deepseek-r1: Incentivizing reasoning capability in llms via reinforcement learning, 2025.
\newblock URL \url{https://arxiv.org/abs/2501.12948}.

\bibitem[Du et~al.(2023)Du, Li, Torralba, Tenenbaum, and Mordatch]{du2023improvingfactualityreasoninglanguage}
Yilun Du, Shuang Li, Antonio Torralba, Joshua~B. Tenenbaum, and Igor Mordatch.
\newblock Improving factuality and reasoning in language models through multiagent debate, 2023.
\newblock URL \url{https://arxiv.org/abs/2305.14325}.

\bibitem[Fu et~al.(2025)Fu, Gao, Shen, Zhu, Mei, He, Xu, Wei, Mei, Wang, Yang, Yuan, and Wu]{fu2025areallargescaleasynchronousreinforcement}
Wei Fu, Jiaxuan Gao, Xujie Shen, Chen Zhu, Zhiyu Mei, Chuyi He, Shusheng Xu, Guo Wei, Jun Mei, Jiashu Wang, Tongkai Yang, Binhang Yuan, and Yi~Wu.
\newblock Areal: A large-scale asynchronous reinforcement learning system for language reasoning, 2025.
\newblock URL \url{https://arxiv.org/abs/2505.24298}.

\bibitem[Gao et~al.(2024)Gao, Xu, Ye, Liu, He, Fu, Mei, Wang, and Wu]{gao2024designingeffectiverlreward}
Jiaxuan Gao, Shusheng Xu, Wenjie Ye, Weilin Liu, Chuyi He, Wei Fu, Zhiyu Mei, Guangju Wang, and Yi~Wu.
\newblock On designing effective rl reward at training time for llm reasoning, 2024.
\newblock URL \url{https://arxiv.org/abs/2410.15115}.

\bibitem[{Gensyn}(2025)]{gensyn2025genrl}
{Gensyn}.
\newblock Introducing rl swarm’s new backend: Genrl.
\newblock \url{https://www.gensyn.ai/articles/genrl}, 2025.
\newblock Accessed: 2025-08-28.

\bibitem[Gensyn(2025)]{gensyn2025rlswarm}
Gensyn.
\newblock Gensyn rl swarm.
\newblock \url{https://github.com/gensyn-ai/rl-swarm}, 2025.
\newblock Accessed: 2025-08-20.

\bibitem[Hessel et~al.(2022)Hessel, Holtzman, Forbes, Bras, and Choi]{hessel2022clipscorereferencefreeevaluationmetric}
Jack Hessel, Ari Holtzman, Maxwell Forbes, Ronan~Le Bras, and Yejin Choi.
\newblock Clipscore: A reference-free evaluation metric for image captioning, 2022.
\newblock URL \url{https://arxiv.org/abs/2104.08718}.

\bibitem[Khan et~al.(2024)Khan, Hughes, Valentine, Ruis, Sachan, Radhakrishnan, Grefenstette, Bowman, Rockt\"{a}schel, and Perez]{khan2024debatingwithmorepersuasivellms}
Akbir Khan, John Hughes, Dan Valentine, Laura Ruis, Kshitij Sachan, Ansh Radhakrishnan, Edward Grefenstette, Samuel~R. Bowman, Tim Rockt\"{a}schel, and Ethan Perez.
\newblock Debating with more persuasive llms leads to more truthful answers.
\newblock In \emph{Proceedings of the 41st International Conference on Machine Learning}, ICML'24. JMLR.org, 2024.

\bibitem[Lambert et~al.(2025)Lambert, Morrison, Pyatkin, Huang, Ivison, Brahman, Miranda, Liu, Dziri, Lyu, Gu, Malik, Graf, Hwang, Yang, Bras, Tafjord, Wilhelm, Soldaini, Smith, Wang, Dasigi, and Hajishirzi]{lambert2025tulu3pushingfrontiers}
Nathan Lambert, Jacob Morrison, Valentina Pyatkin, Shengyi Huang, Hamish Ivison, Faeze Brahman, Lester James~V. Miranda, Alisa Liu, Nouha Dziri, Shane Lyu, Yuling Gu, Saumya Malik, Victoria Graf, Jena~D. Hwang, Jiangjiang Yang, Ronan~Le Bras, Oyvind Tafjord, Chris Wilhelm, Luca Soldaini, Noah~A. Smith, Yizhong Wang, Pradeep Dasigi, and Hannaneh Hajishirzi.
\newblock Tulu 3: Pushing frontiers in open language model post-training, 2025.
\newblock URL \url{https://arxiv.org/abs/2411.15124}.

\bibitem[Le et~al.(2022)Le, Wang, Gotmare, Savarese, and Hoi]{le2022coderlmasteringcodegeneration}
Hung Le, Yue Wang, Akhilesh~Deepak Gotmare, Silvio Savarese, and Steven Chu~Hong Hoi.
\newblock Coderl: Mastering code generation through pretrained models and deep reinforcement learning.
\newblock In S.~Koyejo, S.~Mohamed, A.~Agarwal, D.~Belgrave, K.~Cho, and A.~Oh, editors, \emph{Advances in Neural Information Processing Systems}, volume~35, pages 21314--21328. Curran Associates, Inc., 2022.
\newblock URL \url{https://proceedings.neurips.cc/paper_files/paper/2022/file/8636419dea1aa9fbd25fc4248e702da4-Paper-Conference.pdf}.

\bibitem[Li et~al.(2023)Li, Hammoud, Itani, Khizbullin, and Ghanem]{li2023camelcommunicativeagents}
Guohao Li, Hasan Hammoud, Hani Itani, Dmitrii Khizbullin, and Bernard Ghanem.
\newblock Camel: Communicative agents for "mind" exploration of large language model society.
\newblock In A.~Oh, T.~Naumann, A.~Globerson, K.~Saenko, M.~Hardt, and S.~Levine, editors, \emph{Advances in Neural Information Processing Systems}, volume~36, pages 51991--52008. Curran Associates, Inc., 2023.
\newblock URL \url{https://proceedings.neurips.cc/paper_files/paper/2023/file/a3621ee907def47c1b952ade25c67698-Paper-Conference.pdf}.

\bibitem[Li et~al.(2024)Li, Du, Zhang, Hou, Grabowski, Li, and Ie]{li2024improvingmultiagentdebatesparse}
Yunxuan Li, Yibing Du, Jiageng Zhang, Le~Hou, Peter Grabowski, Yeqing Li, and Eugene Ie.
\newblock Improving multi-agent debate with sparse communication topology, 2024.
\newblock URL \url{https://arxiv.org/abs/2406.11776}.

\bibitem[Liang et~al.(2024)Liang, He, Jiao, Wang, Wang, Wang, Yang, Shi, and Tu]{liang2024encouragingdivergentthinkinglarge}
Tian Liang, Zhiwei He, Wenxiang Jiao, Xing Wang, Yan Wang, Rui Wang, Yujiu Yang, Shuming Shi, and Zhaopeng Tu.
\newblock Encouraging divergent thinking in large language models through multi-agent debate, 2024.
\newblock URL \url{https://arxiv.org/abs/2305.19118}.

\bibitem[Liao et~al.(2025)Liao, Wen, Wang, and Zhang]{liao2025marftmultiagentreinforcementfinetuning}
Junwei Liao, Muning Wen, Jun Wang, and Weinan Zhang.
\newblock Marft: Multi-agent reinforcement fine-tuning, 2025.
\newblock URL \url{https://arxiv.org/abs/2504.16129}.

\bibitem[Liu et~al.(2025)Liu, Liang, Lyu, and Amato]{liu2025llmcollaborationmultiagentreinforcement}
Shuo Liu, Zeyu Liang, Xueguang Lyu, and Christopher Amato.
\newblock Llm collaboration with multi-agent reinforcement learning, 2025.
\newblock URL \url{https://arxiv.org/abs/2508.04652}.

\bibitem[Ma et~al.(2024)Ma, Hu, Pu, Liu, Ai, Liang, and Chen]{ma2024coevolvingwithotheryou}
Hao Ma, Tianyi Hu, Zhiqiang Pu, Boyin Liu, Xiaolin Ai, Yanyan Liang, and Min Chen.
\newblock Coevolving with the other you: Fine-tuning llm with sequential cooperative multi-agent reinforcement learning.
\newblock In A.~Globerson, L.~Mackey, D.~Belgrave, A.~Fan, U.~Paquet, J.~Tomczak, and C.~Zhang, editors, \emph{Advances in Neural Information Processing Systems}, volume~37, pages 15497--15525. Curran Associates, Inc., 2024.
\newblock URL \url{https://proceedings.neurips.cc/paper_files/paper/2024/file/1c2b1c8f7d317719a9ce32dd7386ba35-Paper-Conference.pdf}.

\bibitem[Mistral-AI et~al.(2025)Mistral-AI, :, Rastogi, Jiang, Lo, Berrada, Lample, Rute, Barmentlo, Yadav, Khandelwal, Chandu, Blier, Saulnier, Dinot, Darrin, Gupta, Soletskyi, Vaze, Scao, Wang, Yang, Liu, Sablayrolles, Héliou, Martin, Ehrenberg, Agarwal, Roux, Darcet, Mensch, Bout, Rozière, Monicault, Bamford, Wallenwein, Renaudin, Lanfranchi, Dabert, Mizelle, de~las Casas, Chane-Sane, Fugier, Hanna, Delerce, Guinet, Novikov, Martin, Jaju, Ludziejewski, Chabran, Delignon, Studnia, Amar, Roberts, Denize, Saxena, Jain, Zhao, Martin, Gao, Lavaud, Pellat, Guillaumin, Felardos, Augustin, Seznec, Raghuraman, Duchenne, Wang, von Platen, Saffer, Jacob, Wambergue, Kurylowicz, Muddireddy, Chagniot, Stock, Agrawal, Sauvestre, Delacourt, Gandhi, Subramanian, Dalal, Gandhi, Ghosh, Mishra, Aithal, Antoniak, Schueller, Lavril, Robert, Wang, Lacroix, Nemychnikova, Paltz, Richard, Li, Marshall, Zhang, and Tang]{mistralai2025magistral}
Mistral-AI, :, Abhinav Rastogi, Albert~Q. Jiang, Andy Lo, Gabrielle Berrada, Guillaume Lample, Jason Rute, Joep Barmentlo, Karmesh Yadav, Kartik Khandelwal, Khyathi~Raghavi Chandu, Léonard Blier, Lucile Saulnier, Matthieu Dinot, Maxime Darrin, Neha Gupta, Roman Soletskyi, Sagar Vaze, Teven~Le Scao, Yihan Wang, Adam Yang, Alexander~H. Liu, Alexandre Sablayrolles, Amélie Héliou, Amélie Martin, Andy Ehrenberg, Anmol Agarwal, Antoine Roux, Arthur Darcet, Arthur Mensch, Baptiste Bout, Baptiste Rozière, Baudouin~De Monicault, Chris Bamford, Christian Wallenwein, Christophe Renaudin, Clémence Lanfranchi, Darius Dabert, Devon Mizelle, Diego de~las Casas, Elliot Chane-Sane, Emilien Fugier, Emma~Bou Hanna, Gauthier Delerce, Gauthier Guinet, Georgii Novikov, Guillaume Martin, Himanshu Jaju, Jan Ludziejewski, Jean-Hadrien Chabran, Jean-Malo Delignon, Joachim Studnia, Jonas Amar, Josselin~Somerville Roberts, Julien Denize, Karan Saxena, Kush Jain, Lingxiao Zhao, Louis Martin, Luyu Gao, Lélio~Renard Lavaud, Marie
  Pellat, Mathilde Guillaumin, Mathis Felardos, Maximilian Augustin, Mickaël Seznec, Nikhil Raghuraman, Olivier Duchenne, Patricia Wang, Patrick von Platen, Patryk Saffer, Paul Jacob, Paul Wambergue, Paula Kurylowicz, Pavankumar~Reddy Muddireddy, Philomène Chagniot, Pierre Stock, Pravesh Agrawal, Romain Sauvestre, Rémi Delacourt, Sanchit Gandhi, Sandeep Subramanian, Shashwat Dalal, Siddharth Gandhi, Soham Ghosh, Srijan Mishra, Sumukh Aithal, Szymon Antoniak, Thibault Schueller, Thibaut Lavril, Thomas Robert, Thomas Wang, Timothée Lacroix, Valeriia Nemychnikova, Victor Paltz, Virgile Richard, Wen-Ding Li, William Marshall, Xuanyu Zhang, and Yunhao Tang.
\newblock Magistral, 2025.
\newblock URL \url{https://arxiv.org/abs/2506.10910}.

\bibitem[Motwani et~al.(2025)Motwani, Smith, Das, Rafailov, Laptev, Torr, Pizzati, Clark, and de~Witt]{motwani2025maltimprovingreasoningmultiagent}
Sumeet~Ramesh Motwani, Chandler Smith, Rocktim~Jyoti Das, Rafael Rafailov, Ivan Laptev, Philip H.~S. Torr, Fabio Pizzati, Ronald Clark, and Christian~Schroeder de~Witt.
\newblock Malt: Improving reasoning with multi-agent llm training, 2025.
\newblock URL \url{https://arxiv.org/abs/2412.01928}.

\bibitem[Nguyen et~al.(2024)Nguyen, Shen, Aponte, Xia, Basu, Hu, Chen, Parmar, Kunapuli, Barrow, Wu, Singh, Wang, Gu, Dernoncourt, Ahmed, Lipka, Zhang, Chen, Yu, Kim, Deilamsalehy, Park, Rimer, Zhang, Yang, Rossi, and Nguyen]{vannguyen2024surveysmalllanguagemodels}
Chien~Van Nguyen, Xuan Shen, Ryan Aponte, Yu~Xia, Samyadeep Basu, Zhengmian Hu, Jian Chen, Mihir Parmar, Sasidhar Kunapuli, Joe Barrow, Junda Wu, Ashish Singh, Yu~Wang, Jiuxiang Gu, Franck Dernoncourt, Nesreen~K. Ahmed, Nedim Lipka, Ruiyi Zhang, Xiang Chen, Tong Yu, Sungchul Kim, Hanieh Deilamsalehy, Namyong Park, Mike Rimer, Zhehao Zhang, Huanrui Yang, Ryan~A. Rossi, and Thien~Huu Nguyen.
\newblock A survey of small language models, 2024.
\newblock URL \url{https://arxiv.org/abs/2410.20011}.

\bibitem[OpenAI(2022)]{openai2022instructionfollowing}
OpenAI.
\newblock Aligning language models to follow instructions.
\newblock \url{https://openai.com/index/instruction-following/}, 2022.
\newblock Accessed: 2025-08-19.

\bibitem[OpenAI(2024)]{openai2024learningtoreason}
OpenAI.
\newblock Learning to reason with llms.
\newblock \url{https://openai.com/index/learning-to-reason-with-llms/}, 2024.
\newblock Accessed: 2025-08-19.

\bibitem[Ouyang et~al.(2022)Ouyang, Wu, Jiang, Almeida, Wainwright, Mishkin, Zhang, Agarwal, Slama, Ray, Schulman, Hilton, Kelton, Miller, Simens, Askell, Welinder, Christiano, Leike, and Lowe]{ouyang2022traininglanguagemodelsfollow}
Long Ouyang, Jeff Wu, Xu~Jiang, Diogo Almeida, Carroll~L. Wainwright, Pamela Mishkin, Chong Zhang, Sandhini Agarwal, Katarina Slama, Alex Ray, John Schulman, Jacob Hilton, Fraser Kelton, Luke Miller, Maddie Simens, Amanda Askell, Peter Welinder, Paul Christiano, Jan Leike, and Ryan Lowe.
\newblock Training language models to follow instructions with human feedback, 2022.
\newblock URL \url{https://arxiv.org/abs/2203.02155}.

\bibitem[Park et~al.(2025)Park, Han, Guo, Ozdaglar, Zhang, and Kim]{park2025maporlmultiagentpostcotrainingcollaborative}
Chanwoo Park, Seungju Han, Xingzhi Guo, Asuman Ozdaglar, Kaiqing Zhang, and Joo-Kyung Kim.
\newblock Maporl: Multi-agent post-co-training for collaborative large language models with reinforcement learning, 2025.
\newblock URL \url{https://arxiv.org/abs/2502.18439}.

\bibitem[Perez et~al.(2022)Perez, Huang, Song, Cai, Ring, Aslanides, Glaese, McAleese, and Irving]{perez2022redteaminglanguagemodels}
Ethan Perez, Saffron Huang, Francis Song, Trevor Cai, Roman Ring, John Aslanides, Amelia Glaese, Nat McAleese, and Geoffrey Irving.
\newblock Red teaming language models with language models, 2022.
\newblock URL \url{https://arxiv.org/abs/2202.03286}.

\bibitem[{Qwen~Team}(2024)]{qwen2.5}
{Qwen~Team}.
\newblock Qwen2.5: A party of foundation models, September 2024.
\newblock URL \url{https://qwenlm.github.io/blog/qwen2.5/}.

\bibitem[Schulman et~al.(2017)Schulman, Wolski, Dhariwal, Radford, and Klimov]{schulman2017proximalpolicyoptimizationalgorithms}
John Schulman, Filip Wolski, Prafulla Dhariwal, Alec Radford, and Oleg Klimov.
\newblock Proximal policy optimization algorithms, 2017.
\newblock URL \url{https://arxiv.org/abs/1707.06347}.

\bibitem[Shao et~al.(2024)Shao, Wang, Zhu, Xu, Song, Bi, Zhang, Zhang, Li, Wu, and Guo]{shao2024deepseekmathpushinglimitsmathematical}
Zhihong Shao, Peiyi Wang, Qihao Zhu, Runxin Xu, Junxiao Song, Xiao Bi, Haowei Zhang, Mingchuan Zhang, Y.~K. Li, Y.~Wu, and Daya Guo.
\newblock Deepseekmath: Pushing the limits of mathematical reasoning in open language models, 2024.
\newblock URL \url{https://arxiv.org/abs/2402.03300}.

\bibitem[Stojanovski et~al.(2025)Stojanovski, Stanley, Sharratt, Jones, Adefioye, Kaddour, and Köpf]{stojanovski2025reasoninggymreasoningenvironments}
Zafir Stojanovski, Oliver Stanley, Joe Sharratt, Richard Jones, Abdulhakeem Adefioye, Jean Kaddour, and Andreas Köpf.
\newblock Reasoning gym: Reasoning environments for reinforcement learning with verifiable rewards, 2025.
\newblock URL \url{https://arxiv.org/abs/2505.24760}.

\bibitem[Subramaniam et~al.(2025)Subramaniam, Du, Tenenbaum, Torralba, Li, and Mordatch]{subramaniam2025multiagentfinetuningselfimprovement}
Vighnesh Subramaniam, Yilun Du, Joshua~B. Tenenbaum, Antonio Torralba, Shuang Li, and Igor Mordatch.
\newblock Multiagent finetuning: Self improvement with diverse reasoning chains, 2025.
\newblock URL \url{https://arxiv.org/abs/2501.05707}.

\bibitem[Tian et~al.(2023)Tian, Mitchell, Yao, Manning, and Finn]{tian2023finetuning}
Katherine Tian, Eric Mitchell, Huaxiu Yao, Christopher Manning, and Chelsea Finn.
\newblock Fine-tuning language models for factuality.
\newblock In \emph{NeurIPS 2023 Workshop on Instruction Tuning and Instruction Following}, 2023.
\newblock URL \url{https://openreview.net/forum?id=kEK08VdSO5}.

\bibitem[Wu et~al.(2025)Wu, Wang, Tang, Ding, Helenowski, Tan, Xu, Gowda, Chen, Zhu, Tang, Qian, Zhu, and Hou]{wu2025llamarldistributedasynchronousreinforcement}
Bo~Wu, Sid Wang, Yunhao Tang, Jia Ding, Eryk Helenowski, Liang Tan, Tengyu Xu, Tushar Gowda, Zhengxing Chen, Chen Zhu, Xiaocheng Tang, Yundi Qian, Beibei Zhu, and Rui Hou.
\newblock Llamarl: A distributed asynchronous reinforcement learning framework for efficient large-scale llm training, 2025.
\newblock URL \url{https://arxiv.org/abs/2505.24034}.

\bibitem[Wu et~al.(2023)Wu, Bansal, Zhang, Wu, Li, Zhu, Jiang, Zhang, Zhang, Liu, Awadallah, White, Burger, and Wang]{wu2023autogenenablingnextgenllm}
Qingyun Wu, Gagan Bansal, Jieyu Zhang, Yiran Wu, Beibin Li, Erkang Zhu, Li~Jiang, Xiaoyun Zhang, Shaokun Zhang, Jiale Liu, Ahmed~Hassan Awadallah, Ryen~W White, Doug Burger, and Chi Wang.
\newblock Autogen: Enabling next-gen llm applications via multi-agent conversation, 2023.
\newblock URL \url{https://arxiv.org/abs/2308.08155}.

\bibitem[Yu et~al.(2025)Yu, Zhang, Zhu, Yuan, Zuo, Yue, Dai, Fan, Liu, Liu, Liu, Lin, Lin, Ma, Sheng, Tong, Zhang, Zhang, Zhang, Zhu, Zhu, Chen, Chen, Wang, Yu, Song, Wei, Zhou, Liu, Ma, Zhang, Yan, Qiao, Wu, and Wang]{yu2025dapoopensourcellmreinforcement}
Qiying Yu, Zheng Zhang, Ruofei Zhu, Yufeng Yuan, Xiaochen Zuo, Yu~Yue, Weinan Dai, Tiantian Fan, Gaohong Liu, Lingjun Liu, Xin Liu, Haibin Lin, Zhiqi Lin, Bole Ma, Guangming Sheng, Yuxuan Tong, Chi Zhang, Mofan Zhang, Wang Zhang, Hang Zhu, Jinhua Zhu, Jiaze Chen, Jiangjie Chen, Chengyi Wang, Hongli Yu, Yuxuan Song, Xiangpeng Wei, Hao Zhou, Jingjing Liu, Wei-Ying Ma, Ya-Qin Zhang, Lin Yan, Mu~Qiao, Yonghui Wu, and Mingxuan Wang.
\newblock Dapo: An open-source llm reinforcement learning system at scale, 2025.
\newblock URL \url{https://arxiv.org/abs/2503.14476}.

\bibitem[Yue et~al.(2025)Yue, Yuan, Yu, Zuo, Zhu, Xu, Chen, Wang, Fan, Du, Wei, Yu, Liu, Liu, Liu, Lin, Lin, Ma, Zhang, Zhang, Zhang, Zhu, Zhang, Liu, Wang, Wu, and Yan]{yue2025vapoefficientreliablereinforcement}
Yu~Yue, Yufeng Yuan, Qiying Yu, Xiaochen Zuo, Ruofei Zhu, Wenyuan Xu, Jiaze Chen, Chengyi Wang, TianTian Fan, Zhengyin Du, Xiangpeng Wei, Xiangyu Yu, Gaohong Liu, Juncai Liu, Lingjun Liu, Haibin Lin, Zhiqi Lin, Bole Ma, Chi Zhang, Mofan Zhang, Wang Zhang, Hang Zhu, Ru~Zhang, Xin Liu, Mingxuan Wang, Yonghui Wu, and Lin Yan.
\newblock Vapo: Efficient and reliable reinforcement learning for advanced reasoning tasks, 2025.
\newblock URL \url{https://arxiv.org/abs/2504.05118}.

\bibitem[Zhang et~al.(2025)Zhang, Hosseini, Bansal, Kazemi, Kumar, and Agarwal]{zhang2025generativeverifiersrewardmodeling}
Lunjun Zhang, Arian Hosseini, Hritik Bansal, Mehran Kazemi, Aviral Kumar, and Rishabh Agarwal.
\newblock Generative verifiers: Reward modeling as next-token prediction, 2025.
\newblock URL \url{https://arxiv.org/abs/2408.15240}.

\bibitem[Zhao et~al.(2025)Zhao, Yuksekgonul, Wu, and Zou]{zhao2025siriusselfimprovingmultiagentsystems}
Wanjia Zhao, Mert Yuksekgonul, Shirley Wu, and James Zou.
\newblock Sirius: Self-improving multi-agent systems via bootstrapped reasoning, 2025.
\newblock URL \url{https://arxiv.org/abs/2502.04780}.

\bibitem[Ziegler et~al.(2020)Ziegler, Stiennon, Wu, Brown, Radford, Amodei, Christiano, and Irving]{ziegler2020finetuninglanguagemodelshuman}
Daniel~M. Ziegler, Nisan Stiennon, Jeffrey Wu, Tom~B. Brown, Alec Radford, Dario Amodei, Paul Christiano, and Geoffrey Irving.
\newblock Fine-tuning language models from human preferences, 2020.
\newblock URL \url{https://arxiv.org/abs/1909.08593}.

\end{thebibliography}
\bibliographystyle{plainnat}

\end{document}